\documentclass{amia}
\usepackage{graphicx}
\usepackage[labelfont=bf]{caption}
\usepackage[superscript,nomove]{cite}
\usepackage{color}

\begin{document}

\title{Artificial Intelligence Decision Support for Medical Triage}

\author{Chiara Marchiori, PhD$^{1}$, Douglas Dykeman, PhD$^{1}$, Ivan Girardi, PhD$^{1}$, Adam Ivankay, MS$^{1}$, Kevin Thandiackal, MS$^{1}$, Mario Zusag, MS$^{1}$, Andrea Giovannini, PhD$^{1}$, Daniel Karpati, MS$^{2}$, Henri Saenz, MS$^{2}$}

\institutes{
    $^1$IBM Research, Rueschlikon, Switzerland\\
    $^2$Medgate AG, Basel, Switzerland
}

\maketitle

\noindent{\bf Abstract}

\textit{Applying state-of-the-art machine learning and natural language processing on approximately one million of teleconsultation records, we developed a triage system, now certified and in use at the largest European telemedicine provider. The system evaluates care alternatives through interactions with patients via a mobile application. Reasoning on an initial set of provided symptoms, the triage application generates AI-powered, personalized questions to better characterize the problem and recommends the most appropriate point of care and time frame for a consultation. The underlying technology was developed to meet the needs for performance, transparency, user acceptance and ease of use, central aspects to the adoption of AI-based decision support systems. Providing such remote guidance at the beginning of the chain of care has significant potential for improving cost efficiency, patient experience and outcomes. Being remote, always available and highly scalable, this service is fundamental in high demand situations, such as the current COVID-19 outbreak.}

\section*{Introduction}
Shortage of physicians and increasing healthcare costs have created a need for digital solutions to better optimize medical resources. In addition, patient expectations for mobile, fast and easy 24/7 access to doctors and health services drive the development of patient-centered solutions. The need to be triaged, diagnosed and treated remotely or at home without having to wait in crowded rooms has never been more relevant than in outbreak periods as the COVID-19 one\cite{COVID1,COVID2,COVID3}. Symptom checkers and differential diagnosis generators (DDX) developed for the public in form of web and mobile apps are at the crossroad of these needs. Already several of these developed over the past few years are established on the market. Isabel Healthcare is mostly designed for medical doctors (MDs). It allows the user to enter some demographic information and symptoms and simply suggests a list of all potential diagnoses, ranking them by likelihood and red flagging the potentially dangerous ones. It also displays many literature and web resources for a further exploration of the potential diagnosis (from popular Wikipedia to specialized medical sources such as UptoDate). ADA is developed for the patient. It has a similar starting point, but then engages in a long dialog with the patient to collect additional information on the initially entered symptom and the presence of additional ones. At the end of the dialog, ADA provides a generic and high-level summary on the urgency of the condition and returns a list of the most probable diagnoses (such as ``n out of 10 people with the same symptoms had this condition"). For each displayed diagnosis there is an associated recommendation, such as ``this condition can usually be managed at home", ``seek medical advice", ``seek emergency care".

\begin{figure}[h!]
\centering
\includegraphics[scale=0.25]{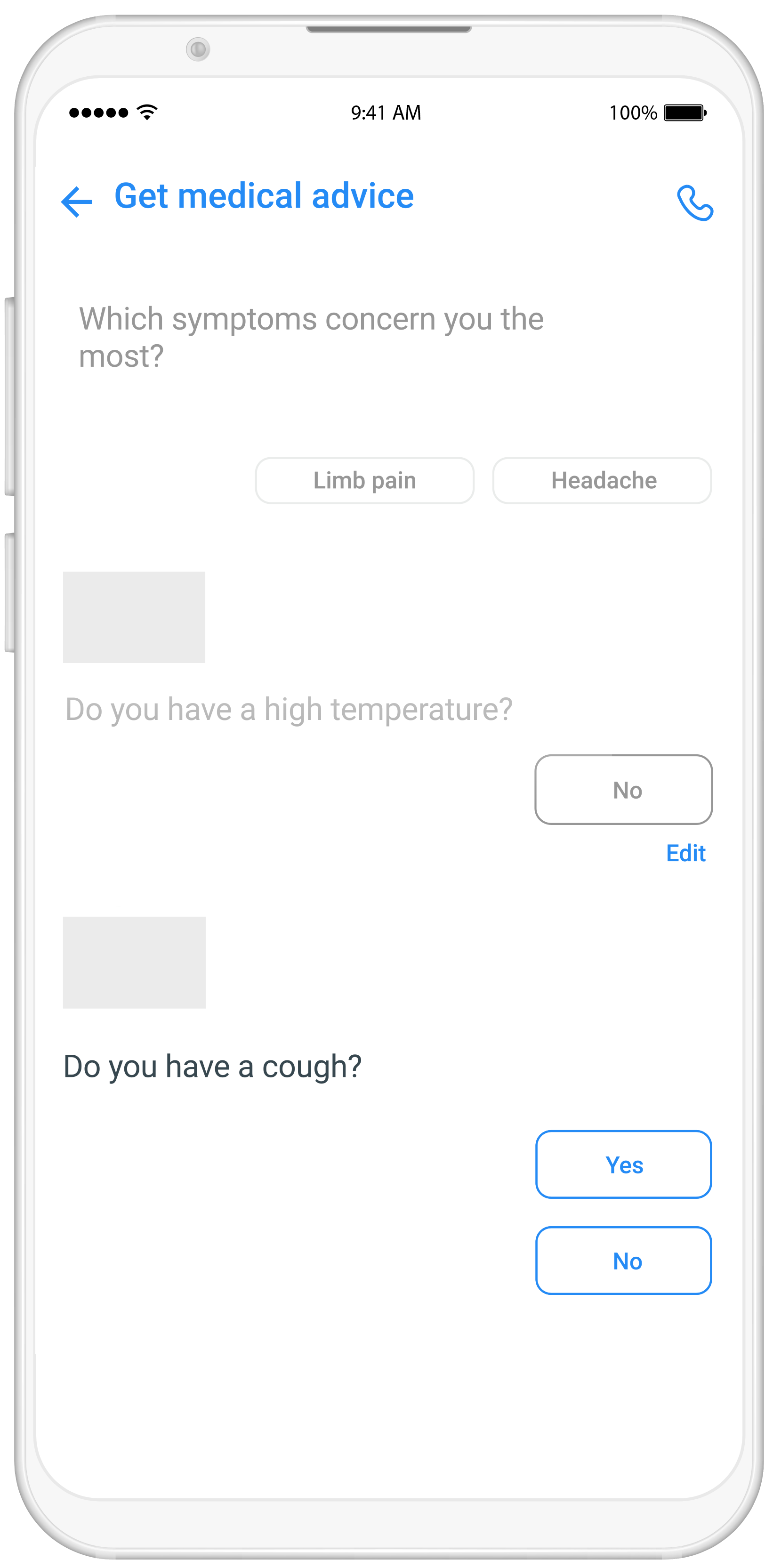}
\hspace{0.2cm}
\includegraphics[scale=0.25]{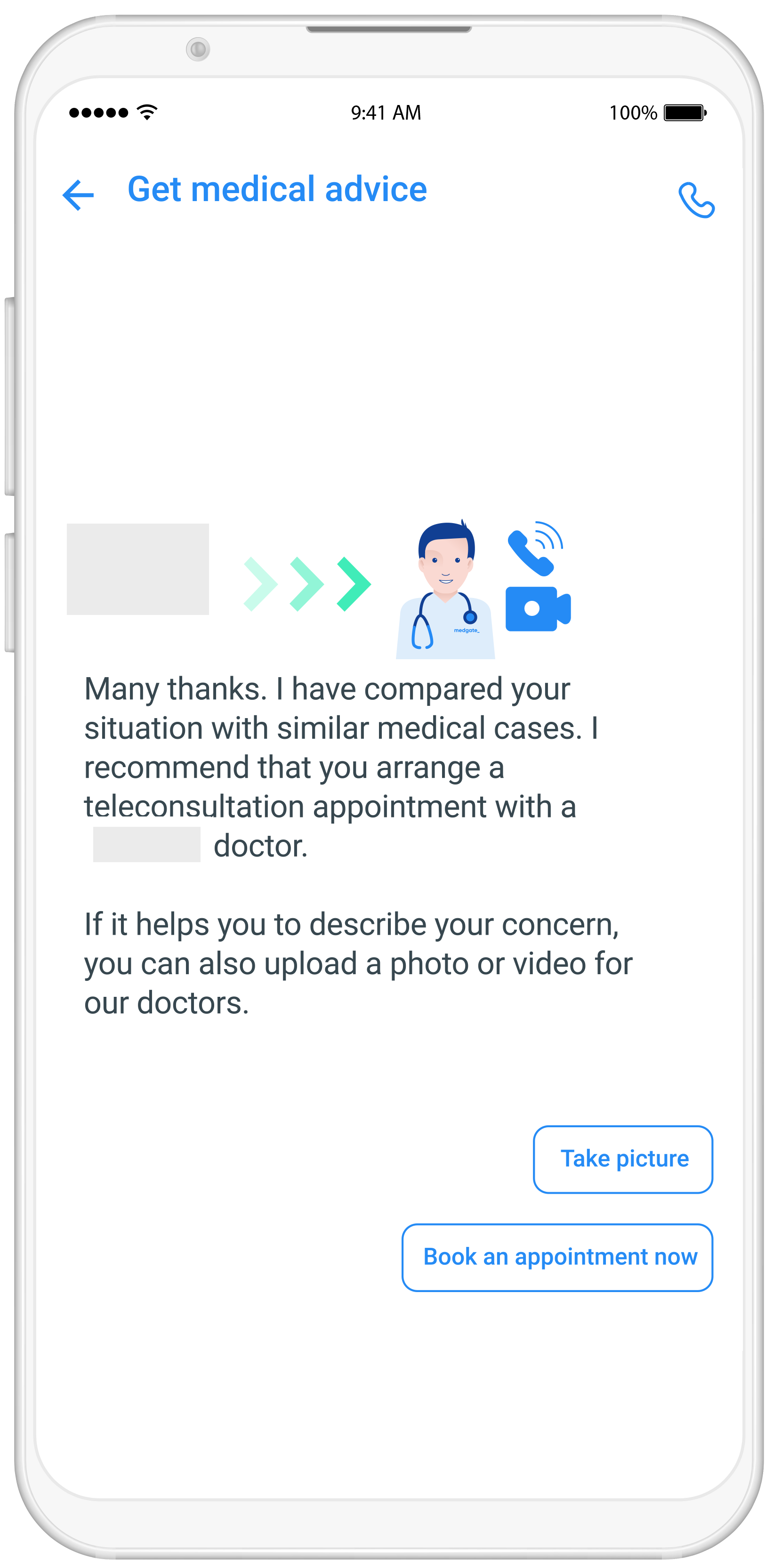}
\hspace{0.2cm}
\includegraphics[scale=0.25]{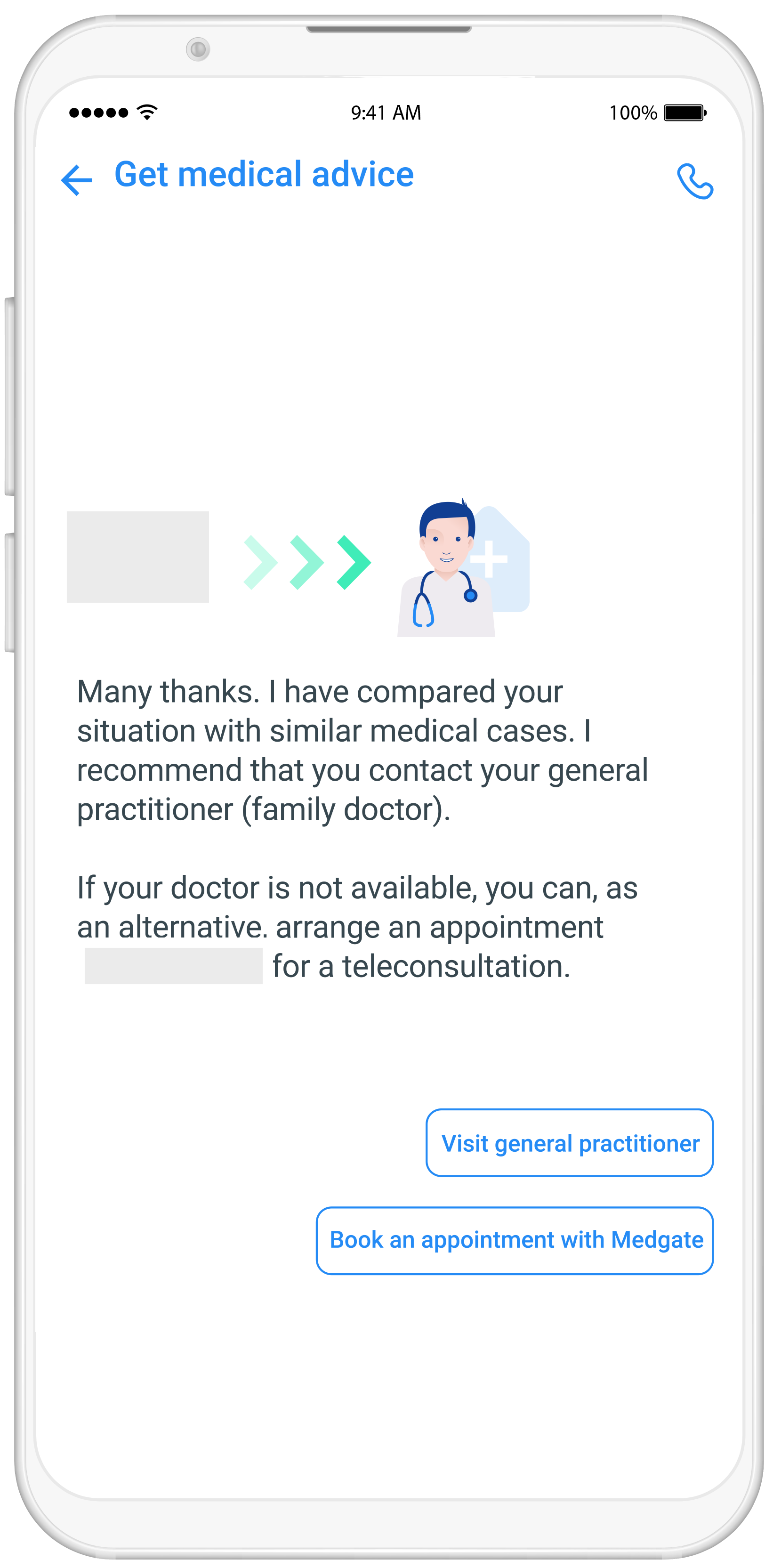}
\caption{Screenshoots of the medical application currently in use.}
\label{Fig:app_screenshots}
\vspace{0.2cm}
\end{figure}

\begin{figure}[h!]
\centering
\includegraphics[scale=0.3]{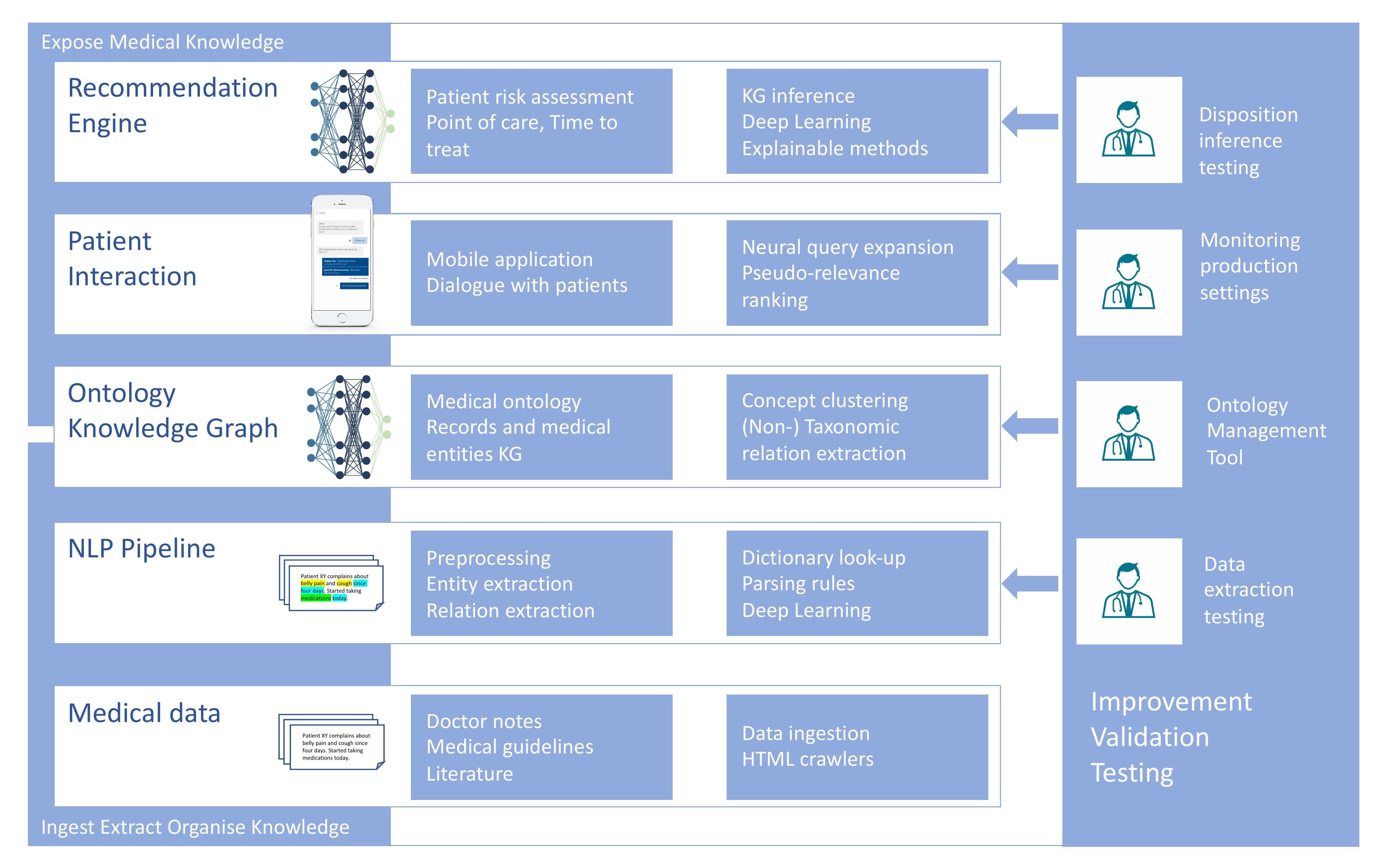}
\caption{Overview of the solution building blocks. Right column: improvement, testing and validation tasks performed with physicians help.}
\label{Fig:pipeline-stages}
\vspace{0.2cm}
\end{figure}

The telemedicine triage that we developed, the Artificial Intelligence Triage Engine (AITE), is an integral part of a rich ecosystem of telemedicine services and is fundamentally different from symptom checkers in several aspects. Starting with few input symptoms (see Figure~\ref{Fig:app_screenshots}), AITE engages in short and focused dialogues to finally provide clear and simple recommendations to patients. Patients receive suggestions on which medical service provider to go to depending on the characteristics of their condition. In critical cases, patients are recommended to call the telecare provider immediately for a teleconsultation. If the health condition of the patient is assessed as being less urgent, they can select an appointment of their choice via a phone or video consultation with a trained telecare specialist (see Figure~\ref{Fig:pipeline-stages}). Should patients require physical intervention, they will be referred to an appropriate care provider for a face-to-face consultation. The advantages are clear. By using the mobile application, emergency cases can be treated faster than in standard call center settings. Moreover by automatically filtering out patient requiring physical intervention, telemedical resources can invest more time in treating telecare eligible patients.
Furthermore, to recommend the most appropriate course of action, the app draws from the patient's personal medical history as well as from similar cases handled by the telecare provider in the past.

Built on case records and guidelines using AI-based methods, the system consists of the following building blocks: 1) an engine for the automated ingestion of unstructured clinical notes, the extraction of relevant medical entities and their organization into a knowledge graph (KG); 2) a data-driven dialog system that allows a conversation with such medical knowledge base and drives the patient interactions; 3) an inference engine able to suggest the most appropriate recommendation in terms of point of care and time frame for treatment. All these components are depicted in Figure~\ref{Fig:pipeline-stages} and will be discussed in detail in the following sections.

\section*{Data}
To build the triage application described here, we used more than 900k case records written in German and collected over more than 7 years. This is only a fraction of the available data, since only records generated by top-ranked doctors (based on years of experience and internal audits) were selected. The records are notes that call center agents and doctors took while talking to the patients over the phone. They are structured in sections that contain demographic data such as age and gender, previous illnesses, and free text descriptions of the patient's current medical condition. It should be noted that in this kind of records, the free text contains a shallow and subjective description of the patient's problem. Patient narrative can contain mention of recent surgery or treatments and whether the patient can perform or not a specific test/movement, but no physical exams nor lab tests. Potential diagnoses consistent with the symptom description are listed. The descriptions in the records are expressed in formal medical language as well as in layman's terminology. Typical for these settings, sentences are not always complete (e.g., subjects or verbs may be missing) and include misspellings, dialect vocabulary, non-standard medical abbreviations and inconsistent punctuation. This is a challenge for the linguistic processing of the case records.

\section*{NLP pipeline}
A natural language processing (NLP) pipeline extracted medically relevant concepts from each case record. The pipeline consisted of the following stages: (1) data preprocessing for misspelling correction and abbreviation expansion, (2) named entity recognition (NER) and (3) concept clustering for the dynamic creation of an ontology of medical concepts from the corpus. Acronyms and abbreviations used unambiguously were linked to the corresponding entities directly in the medical dictionaries. Ambiguous acronyms and abbreviations were resolved, when possible, using algorithms that include context for disambiguation. Although we were able to detect many entity types, i.e. from anatomy and physiology to medical procedures and medicines, the main focus was on the extraction of current, non-current and negative mentions of symptoms and diagnosis in simple and complex expressions. Special attention was devoted to symptom and disease characterization in terms of body part location (e.g. \textit{pain in the leg}, \textit{abscess in the arm}), intensity (\textit{strong}, \textit{light}), time duration (\textit{chronic}, \textit{acute}), activity by which the symptom manifests or changes in character (e.g. \textit{dyspnea at rest}, \textit{exertional dyspnea}). For medical entity extraction we used a system based on the combination of dictionary look-up, advanced rules and deep learning. The dictionaries used in the NER were built partly using existing German-language medical dictionaries and ontologies (UMLS mapped German terms, ICD10, MedDRA, etc.) and partly using the list of words contained in the case records. They therefore contain a mapping of technical and layman's terms, for a total of more than 140k unique words (including declination, capitalization). Negated mentions of medical entities are very frequent in this type of records and were detected using German language-specific negation particles or expressions.

For relations that associate a medical entity to a specific body part, advanced parsing rules were used to detect short distance relations with high precision, while a deep learning (DL) stage was used to detect distant relations, improving therefore the overall recall. We trained a DL binary classifier with positive and negative examples. The annotated data consisted of $10k$ triples (E1, R, E2) manually generated on the raw texts; where E1 is a symptom, disease, or operation, E2 is an anatomical location, and R is the positive or negative relation between the two entities. To obtain a good balance of training examples, 50\% triplets had R = ``located\_in" and 50\% triplets had R = ``not\_located\_in". This approach can also be used in case of multi-relation extraction by building a binary classifier for every relation, where the use of correctly balanced negated examples helps improve the training.
The raw text data was processed by (a) removing stop words, (b) transforming to lower case, (c) normalizing to ASCII characters, (d) transforming delimiters (e.g. ``;", ``,", ``.") to word embeddings by adding white spaces, and (e) excluding any non-alphanumerical characters and characters not in the delimiter set. For each word, we concatenated word-, positional- and part-of-speech-features (POS-features) creating the input feature vector. Word features were constructed using pre-trained word embeddings on the raw text data, using either word2vec\cite{word2vec} or ELMo2 \cite{ELMo} embeddings. Positional features were defined as the combination of the relative distances of the current word from the entities E1 and E2\cite{Zeng2014}. 
We explored three different architectures Convolutional Neural Network\cite{Kim2014} (CNN) and Bidirectional Gated Recurrent Unit Network (Bi-GRU), Bidirectional Long Short Term Memory Network (Bi-LSTM), combining the models proposed in \cite{RNN,LSTM,BRNN,GRU1}.

\begin{figure}[h!]
\centering
\includegraphics[scale=0.25]{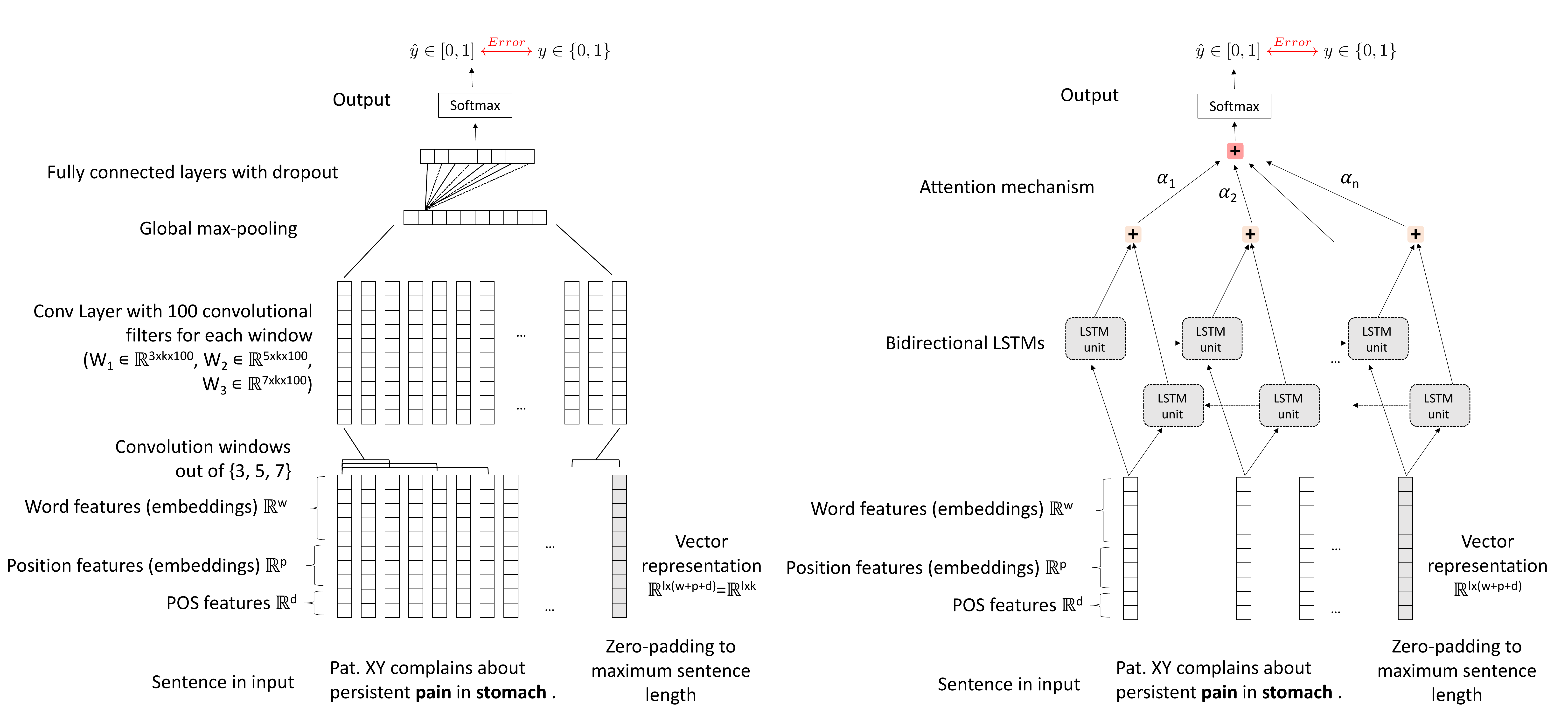}
\caption{Architecture of the CNN model (left panel) and Bi-LSTM model (right panel) for relation extraction.
In the figure $w$, $p$, $d$ are the dimensions of word-, positional and POS-features, $l=45$ is the maximum sentence length.}
\label{Fig:NN}
\vspace{0.2cm}
\end{figure}

In the CNN architecture, the input feature vector with dimension $k$ was fed to convolutional layers. These layers encoded the word sequence into $n$-gram representations, to capture the contextual information. For a given sentence, a weight matrix $w \in R^{m \times k}$ was applied to generate the new features $c_i$ from the window of words $x_{i:i+m-1}$ where:
\begin{equation}
    c_{i} = Relu(w\cdot x_{i:i+m-1} + b)\,,\newline
\end{equation}
with $m$ convolutional window.
This filter was applied to each possible window of words in the sentence
\begin{equation}
{x_{1:m},x_{2:m+1},.....x_{n-m+1:n}}
\end{equation}
to produce a feature map:
\begin{equation}
    c = [c_{1},c_{2},......,c_{n-m+1}],
\end{equation}
\noindent with \(c \in R^{n-m+1}\).
By applying multiple filters (denoted \(f\)) on \(x_{i:i+m-1}\), we obtained a new representation of the sentence. By setting different values for $m$, we obtained different $n$-gram representations of the sentence which was input to a max pooling layer, similarly to \cite{Zeng2014, Collobert2011, Kim2014}.
The last layers of the CNN architecture comprised several fully connected layers with dropout and a softmax for the classification. The architecture schema of the CNN model is shown in the left panel of Figure~\ref{Fig:NN}. The dimensions of the word embeddings $w$, the positional embeddings $p$, the number and size of the convolutional windows, the number of filters $f$, the number of fully connected layers and their dimensions, the number of epochs and the dropout rate were tuned during training. In the Bi-GRU and Bi-LSTM architectures, the input feature vector with dimension $k$ was fed into Bi-GRU or Bi-LSTM layers, followed by an attention and a softmax layer for the classification. The architecture schema of the Bi-LSTM is given in the right panel Figure~\ref{Fig:NN}.

For all architectures, we used grid search to find the optimal values of the model-specific hyperparameters. We used a randomly constructed cross-entropy loss function with mini-batch updating and Adam optimizer for five epochs. In our experiment, we found optimal values for $w=50$ and $p=20$.
The annotated data was split in 60\% training, 10\% validation and 30\% testing.
Precision, recall and f-score on the test set on the different architectures are shown in Table~\ref{table1}.
\begin{table}[h!]
\centering
\vspace{0.5cm}
\caption{Prediction results of the different architectures on relation extraction (i.e., predicting whether the relation is $R_1$ or $R_2$), where P($R_k$), R($R_k$), F($R_k$) are precision,
recall and f-score and $R_1 =$ ``not\_located\_in",
$R_2 =$ ``located\_in".
Similar values were obtained by conducting several experiments and averaging the results.}
  \begin{tabular}{|l|l|l|l|l|l|l|}
  \hline
    \textbf{Experiment}    & \textbf{P($R_1$)}  & \textbf{R($R_1$)} & \textbf{F($R_1$)} & 
    \textbf{P($R_2$)}  & \textbf{R($R_2$)} & \textbf{F($R_2$)} 
    \\ \hline
    Word2Vec-CNN  & 0.77 & 0.84 & 0.81 & 0.82 & 0.74 & 0.77  \\ \hline
    Word2Vec-Bi-GRU  & 0.90 & 0.88 & 0.89 & 0.77 & 0.80 & 0.78\\ \hline
    ELMo-Bi-LSTM  & 0.92 & 0.90 & 0.91 & 0.80 & 0.85 & 0.83 \\ \hline
  \end{tabular}
\label{table1}
\end{table}

Notably, the described NLP pipeline is able to ingest one million patient records at run time and extract the relevant medical entities and their relations in about two hours. Therefore, if needed, new, improved versions of the system could be produced almost on a daily basis, by ingesting new case records whenever available. A highly performant pipeline is also important for efficient debugging of the NLP algorithms.

\section*{Ontology and Knowledge Graph}
Several approaches have been developed for ontology learning\cite{Biemann2005, Wong2012}, which is defined as the extraction of terms, concepts, taxonomic relations and non-taxonomic relations from data. An ontology can be built from scratch for example using clustering algorithms or from an existing ontology performing classification tasks.

The first step in the creation of the ontology in our application was the grouping of the annotations gathered through the NLP pipeline that describe the same medical concept. For this process we used a hierarchical procedure, named concept clustering. This is important to improve  performance of the inference engine to ensure high recall for systems based on patient similarity and reduce feature correlation for
machine learning classifiers.
It is especially laborious in German, due to the frequent use of complex compound names that can be also expressed by several permutations of the corresponding simple entities. For example, \textit{Augendruckschmerz} - painful sensation of pressure in the eye - contains the three simple entities [\textit{Auge}, \textit{Druck}, \textit{Schmerz}] -- [\textit{eye}, \textit{pressure}, \textit{pain}] -- and can also be expressed by \textit{Druckschmerz am Auge}, \textit{schmerzhaftes Druckgefühl hinter den Augen}, etc. This concept clustering was performed in consequent stages, at dictionary level first, by grouping all synonyms in one single dictionary entry (e.g. abdomen, belly) and at annotation level, by grouping annotations with similar semantic blocks, e.g. [\textit{belly}, \textit{pain}], [\textit{abdominal}, \textit{pain}]. This multi-stage approach can be used to tailor the level of granularity of the medical concepts to the different applications and end users, in our case to medical triage and patients. For example, if such a level of detail is not required, all explicit mentions of \textit{finger phalanxes} can be merged into the concept \textit{finger} at a customized annotation level.

The created medical concepts are then organized in an ontology, where the following relations among medical concepts were considered: 1) child of (Taxonomic Relations) and 2) located in, negation of, characterization of, specification of (Non-Taxonomic Relations). The relations child-of were automatically built using as a reference the relations contained in two manually curated elementary ontologies, ``Anatomy" (e.g. Augenlid is child of Auge - eyelid is child of eye) and ``General Symptom" (e.g. Druckschmerz is child of Schmerz), which were derived from standard medical resources(e.g., \cite{FMA}).
To summarize, more than one million medical expressions were merged into $17k$ medical concepts specific to medical triage, related to a very broad range of problems (e.g. urinary, digestive, respiratory, locomotive, skin, eye). These medical concepts were organized in a large ontology containing $17k$ nodes ($15k$ symptoms, $2k$ diseases) and $100k$ edges. In addition to a name and its synonyms, each medical concept was also characterized by metadata, used, for example, to assign the semantic type (symptoms, disease), distinguish between red flags and common symptoms, female or male specific conditions, psychological or other types of symptoms. Metadata are consumed by the system which drives the dialogue with the end users (Q\&A system).

\begin{figure}[h!]
\centering
\includegraphics[scale=0.45]{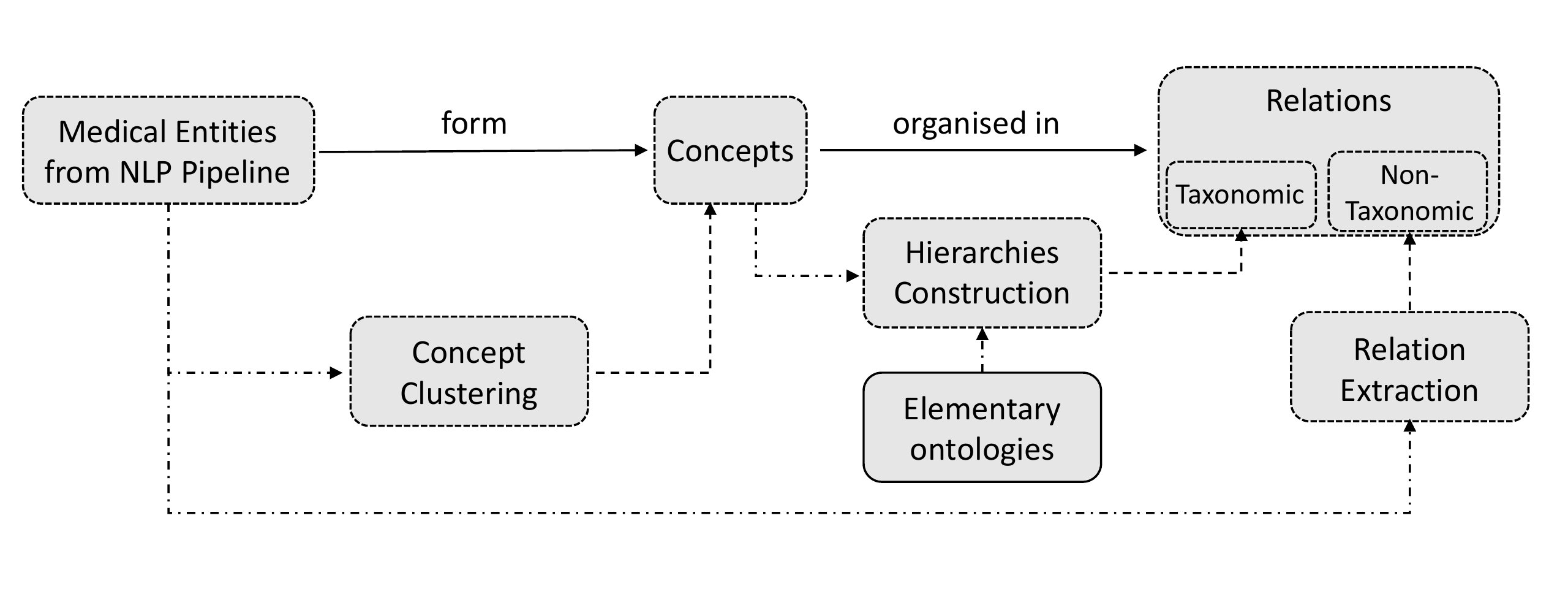}
\caption{Diagram of the stages used for ontology learning and their outputs. The stages are connected to the outputs with dashed arrows.}
\label{Fig:OL_diagram}
\vspace{0.1cm}
\end{figure}

The diagram of the stages used to construct the ontology is summarized in Figure~\ref{Fig:OL_diagram}.

It should be noted, that this solution allowed for the creation of an ontology of medical concepts directly and automatically from the ingested data, whereby layman's and technical terminology as well as many synonym expressions are mapped to the same concept. In addition, this solution is useful for languages that are not yet well covered by standard ontologies such as UMLS/SNOMED.

After ontology creation, the input case records together with the extracted medical concepts and metadata were automatically ingested and organized in a language agnostic knowledge graph (KG). KGs are structured knowledge bases (KBs) that store information in form of relationships (edges) between entities (nodes or vertices)\cite{KG1, KG2, KG3}. The graph is represented by its sparse adjacency matrices, indicating which vertices are connected with a given relation. The created KG consisted in more than 1M of nodes of the type \textit{case record}, \textit{age}, \textit{gender}, \textit{symptom}, \textit{disease}, \textit{red flags}, \textit{historical recommendation} and 23M edges linking these entities, such as, for example \textit{symptom-to-patient}, \textit{disease-to-patient}, \textit{patient-to-point-of-care}, \textit{age-group-to-patient}.
The language agnostic KG was built for multilingual triage applications by first using data in one specific language (German) as a reference and then mapping the medical entities (KG nodes) to numerical codes. To obtain language interoperability, these numerical codes should correspond to international standards, such as the UMLS concept unique identifier, or ICD and SNOMED codes. However, the coverage of these coding systems for the specific languages can vary considerably and therefore a complete mapping was not possible.

\section*{Patient Interaction}
An essential part of the triage process is the Q\&A session, during which additional, carefully chosen symptoms are asked to the patient. As the patients do not have the necessary knowledge for proper triage, this process ensures that they enter all relevant symptoms, even the ones that they might consider irrelevant or unnecessary to input. Users might not know correlations between symptoms that are important.
The question generation algorithm drives the interaction with the patient. Its goal is to identify which medical concepts need to be asked as the most relevant to the initial input provided by the patient, emulating the human expert decision process. In our system, the question generation algorithm is fully learnt from the data: the questions are determined dynamically based on the patient's input and the system's knowledge on the training data. We explored two different approaches. One based on information retrieval algorithms, i.e., pseudo-relevance feedback based query expansion techniques (such as entropy, mutual information) and the second one on neural network techniques. In the first approach question generation and patient risk assessment are jointly optimised so that the collection of additional relevant information from the patient aims to improve the classification task. While in the second approach the two tasks are optimised (i.e., trained) separately. 

For the neural network based model we constructed a training corpus masking one or multiple medical concepts from each patient case and optimised the network to predict the obscured features. We split the data in 70\% for training, 10\% for validation and 20\% for testing. Due to the high number of word features that may be obscured, the prediction task becomes very hard. A random classifier would only achieve an accuracy of $5.6 \times 10^{-5}$ (inverse of the number of word features) for the top one (i.e., highest probability) prediction.

\begin{table}[h]
\centering
\vspace{0.5cm}
\caption{Testing results of a neural network based (NN) query expansion module compared to information retrieval
techniques. In the table we show the top-k accuracy (Acc@k), using inverse probability masking.}
  \begin{tabular}{|l|l|l|l|}
  \hline
    \textbf{Model}    & \textbf{Acc@1}  & \textbf{Acc@5} & \textbf{Acc@10}
    \\ \hline
    f  & 0.0095 & 0.087 & 0.175  \\ \hline
    BIM  & 0.0093 & 0.037 & 0.0578  \\ \hline
    CHI  & 0.0019 & 0.0336 & 0.0664  \\ \hline
    RSV  & 0.0107 & 0.0873 & 0.170  \\ \hline
    KLD  & 0.0016 & 0.0654 & 0.138  \\ \hline
    NN  & 0.13 & 0.26 & 0.34  \\ \hline
  \end{tabular}
\label{table:nbq}
\end{table}

In Table~\ref{table:nbq} we compared the performance of a neural network based query expansion model\cite{NBQPatent2019} with the following rankers: i) frequency-based (f), ii) binary independence model (BIM\cite{BIM}), iii) $\chi^2$ value (CHI\cite{CHI2}), iv) Robertson selection value (RSV\cite{RSV}), v) Kullback-Leibler divergence (KLD\cite{KLD}, relative entropy). For a comprehensive review of those methods see also \cite{IRreview}. 
The frequency-based ranking of the terms (f) ranks the most often occurring symptoms the highest, while low-occurrent ones lower as in the formula $F(s, R) = \sum_{r{\in}R}I_{s{\in}r}\,$,
where $R$ is the set of relevant cases retrieved, $s$ is the token to be ranked and $I$ the identity if the token $s$ belongs to the set $r$, zero otherwise. A drawback of this method is that it ranks potentially irrelevant symptoms high, that are only common because they are frequent in the data, hence also in the relevant subset of cases. The binary independence model (BIM) ranks symptoms based on the equation below:
\begin{equation}
    BIM(s,R,N) = log\frac{p(s|R){\cdot}[1-p(s|N)]}{p(s|N){\cdot}[1-p(s|R)]}\,,
\end{equation}
where $s$ denotes the symptom to be ranked, $R$ is the set of relevant retrieved cases and N is the case base. $p(s|R)$ and $p(s|N)$ stand for the probabilities of the symptom occurring in the relevant set of cases and the whole case base respectively. The exact probabilities are unknown but can be approximated with the counts. The $\chi^2$ value is used to determine whether there is significant difference in expected values of two events. In our case, these two events are the query occurrence in the relevant set of cases and the occurrence in the whole set of training data. The definition is given in the equation below:
\begin{equation}
    CHI(s,R,N) = log\frac{[p(s|R)-p(s|N)]^2}{p(s|N)}\,.
\end{equation}
The Kullback-Leibler divergence (KLD, relative entropy) of the aforementioned probabilities is also used as a term ranking metric. It is defined as written in the equation below: 
\begin{equation}
    KLD(s,R,N) = p(s|R){\cdot}log\frac{p(s|R)}{p(s|N)}\,.
\end{equation}
KLD is a measure of how two probability distributions differ from each other. The Robertson selection value (RSV) ranks the symptoms based on the product of their differences in occurrence probabilities and their term weight for each document in the relevant records. This is intuitively promising as certain tokens might be low-occurrent compared to others, yet of major importance for the downstream classification task, which is captured by a higher term weight. RSV is described in the equation below:
\begin{equation}
    RSV(s,R,N) = \sum_{r{\in}R}w(s, r){\cdot}[p(s|R)-p(s|N)]\,,
\end{equation}
where $w(s, r)$ denotes the weight of term $s$ in the given relevant record $r$. 
NN based question generation systems show very promising results in order to generate questions that are medically relevant and consistent with the initially provided symptoms, hence fundamental for the construction of a Q\&A system. The described experiment has been used to benchmark different algorithms. In real settings, the question generation system is only one component of the dialogue system and only a subset of medical conditions is used. A key aspects in the dialogue is the usage of the medical ontology.
Central to the development and digitization of AI-based decision support system are user acceptance and user friendliness. In AITE, for example, patients can state their symptoms in a natural way since many different forms of expressing the same symptom have been learned from the training data provided by the telemedicine provider. An elastic search engine provides the mapping from the user input to the medical concepts using all variety of symptom expressions extracted from the data and stored in the ontology.
The Q\&A part plays an important role in defining the overall user experience, hence speed and usability are crucial. In order to achieve the best user experience, the Q\&A was customized according to patient age, gender, and medical status, and special attention was paid to a simple and understandable formulation of the questions. Questions should be relevant to the input symptoms and conversations should be short. This required that a final recommendation could be reached in only few interactions, with a short waiting time between these, i.e., less than one second. 
Especially difficult was to design a system to avoid asking ``similar questions" that are interpreted by a human as repetitive and redundant. This was partially solved by using the hierarchical structure of the medical ontology, by excluding all the children of a medical concept as possible question candidates, if the patient has denied having their parent. For instance, the system will not ask any symptoms related to a specific part of the abdomen (upper, lower, right, left, center) if the patient denies having an abdomen problem. These Key Performance Indicators (KPIs) were explicitly evaluated during several testing campaigns in different settings which led to continuous refinement of the Q\&A model and logic. 

\section*{Recommendation engine}
The inference engine is the part of the system that will reason on the collected information and give a final recommendation. The are several approaches to build a recommendation engine for patient risk assessment. 
In one type of approach, patient similarity is calculated on the KG using node and edge weights, to identify the sub-graph region containing the patient data most similar to the user. This region might also be used to retrieve the possible key missing medical conditions, in systems where question generation and recommendation inference are coupled. Node and edge weights in the KG are learned from the data and medical guidelines with cross entropy cost minimisation. In a second type of approach, the KG is used as a knowledge base to extract feature vectors, (e.g. by learning embeddings from the graph\cite{DeepWalk, LINE, node2vec, Yang2016}, combining embeddings learned from the original text with graph features) to build a machine learning classification engine. While it is not possible to report final accuracy levels due to confidentiality, we tested an implementation of this second approach which achieved f-scores as high as 87.5\%, 74.0\% and 90.4\% on high-, medium- and low-risk classes respectively, with a confidence threshold of 0.6. Based on CNN methods \cite{Kim2014}, it is extensively described in\cite{AIDA2018}. Finally, patient risk assessment can be performed with high classification performance also with advanced DL techniques. In this latter case, however, the methods are perceived as a `black box" in terms of how they generate the predictions from the input data and the addition of an explanation method is needed for user acceptance\cite{AIDA2018}.
In implementing the first approach, we modeled the retrieval of similar
patients on the KG with a sequence of graph operations. The technology was developed on proprietary libraries on sparse linear algebra \cite{KG1, KG2}. The final implementation was extremely efficient and allowed for very fast graph operations. The computation of graph traversals with arbitrary depth for sparse adjacency matrices was of the order of milliseconds ($O(ms)$) for single graph traversal. High performance was obtained by keeping the KG in memory. The response time for any single query was measured to be well below the four seconds requirement with a single worker, as the system architecture enables scaling, and thus efficient large-scale patient support. The KG was also used for traceability and transparency in the recommendation rationale. The retrieval of similar patients to the users allows to understand why a certain recommendation was given from historical cases.

\begin{figure}[h!]
\centering
\includegraphics[scale=0.6]{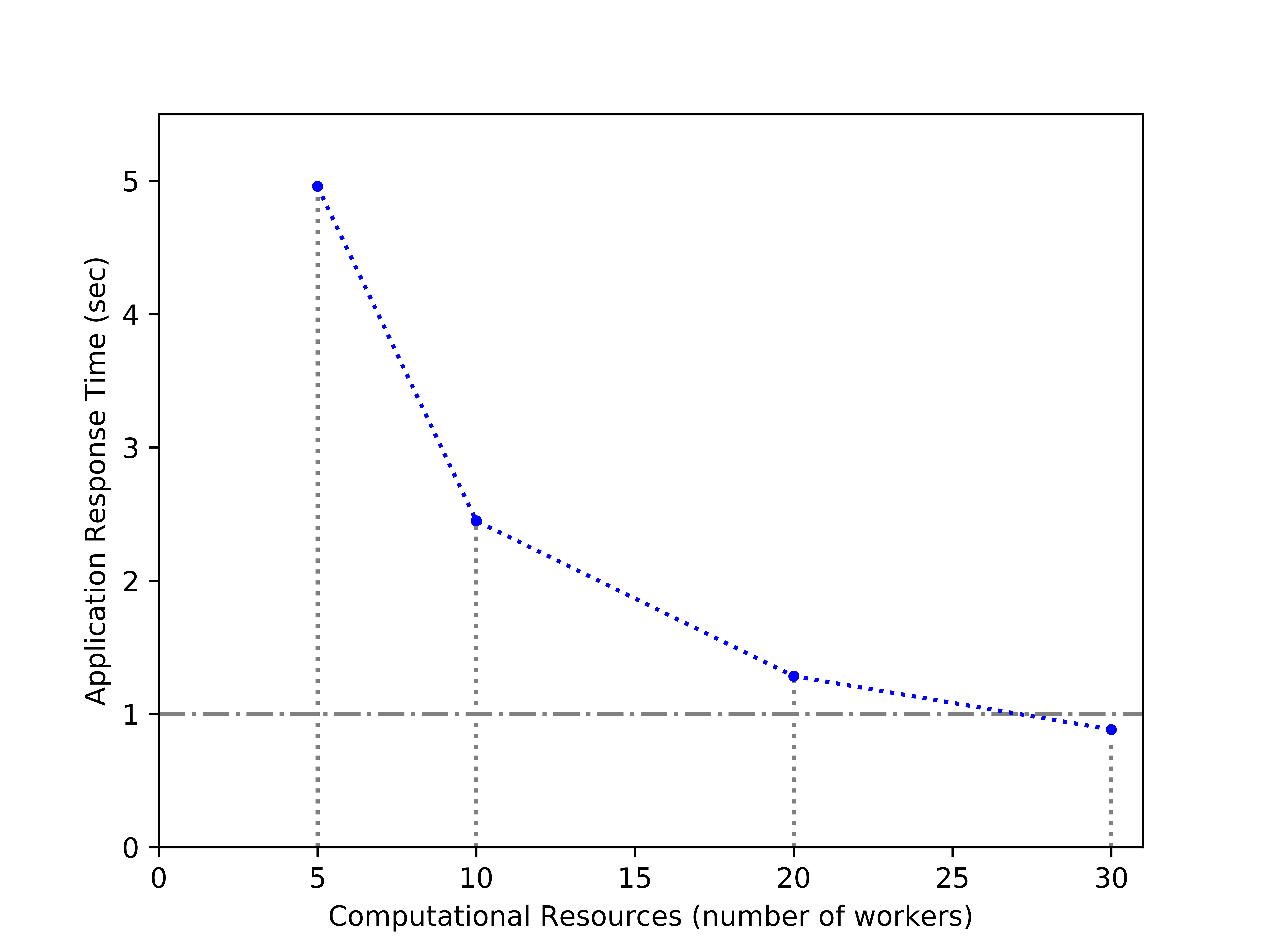}
\caption{Average application response time in seconds for increasing computational resources (number of workers)
with $30$ concurrent user requests.}
\label{Fig:app_perf}
\vspace{0.2cm}
\end{figure}

A key requirement for usability and user-friendliness was speed and scalability. Patient requests can be spread over the available servers, and speedup from concurrency is linear with the available hardware resources as shown in Figure~\ref{Fig:app_perf}. All patient interactions can be handled on average in less than one second with enough computational resources.

\section*{Subject matter expertise in clinical validation and testing}
Clinical testing and validation were fundamental for the certification of the solution. In such a complex system, testing and validation were performed at different levels, encompassing: 1) semi-automated validation of single building blocks (e.g. NLP, Ontology, Q\&A); 2) automated validation of the end-to-end output on a set of clinically validated ground truth cases (\textit{automated recommendation testing}); (3) validation of the system in real-life and production settings. Whenever possible automated testing procedures were streamlined to the execution of the pipeline for the creation of the system, so that they could be run at any update or modification of training corpus or pipeline components. Physicians support was fundamental in all testing steps as depicted in the right column of Figure~\ref{Fig:pipeline-stages}. \textit{Automated recommendation testing} was performed by first creating a ground truth set in which patient demographic, symptoms and the corresponding recommendations were known and validated by clinicians. Recall, sensitivity, precision and f-score were then calculated by comparing those recommendations with the ones given by the system on the same input. Although metric values can not be disclosed, it has to be noted that all had to be above required thresholds (with a special attention to emergency recall) for system release . Validation of the system in real life was performed by physicians and other subject matter experts, by evaluating defined KPIs on a statistically significant sample during the usage of the tool in different settings. To reduce the time and effort requested to the physicians in these activities, special methods and tools were developed to enable a continuous loop between clinician review and quick implementation.

\section*{Conclusion}
 We have presented results, solutions and approaches used to build an interactive, AI-based system for automated medical triage from a large corpus of case records. We have considered various aspects in developing such an end-to-end system, from precision on the recommendation, transparency for trust, adoption and end-user friendliness, to the system scalability in terms of number of users served.
 A key aspect of our research work was the development of a pipeline capable of automatically generating a very rich ontology of medical concepts and a knowledge base, directly from a large corpus of ingested records. This included a highly efficient DL-augmented NLP stage with high precision and recall for the extraction of medical entities and their relations, and a language agnostic implementation of a medical knowledge graph for multilingual applications. A second key aspect was the implementation of reasoning methods on such a knowledge base, comprising a user-friendly, fully data-driven question generation technology and an inference engine for the final recommendation. A third key aspect was that several testing procedures were automated and streamlined to the development activities, for a continuous and consistent feedback implementation. The abstraction and modularity of the underlined solutions are crucial for the reusability of the components in general settings such as automated knowledge ingestion and organisation, development of dialogue systems and decision support applications.
  Providing such automated triage services will help healthcare providers satisfy a larger number of patients and focus their valuable medical resources on telecare eligible patients at the same time. The optimisation of the medical resources will improve patient experience overall.

\makeatletter
\renewcommand{\@biblabel}[1]{\hfill #1.}
\makeatother

\bibliographystyle{unsrt}
\centering

\end{document}